\documentclass[sigconf,natbib=true,anonymous=false]{acmart}
\usepackage[ruled,linesnumbered]{algorithm2e}
\usepackage{multirow}
\usepackage{makecell}
\usepackage{enumitem}
\usepackage{diagbox}
\setitemize[1]{noitemsep,partopsep=0pt,parsep=0pt,topsep=0pt, leftmargin=10pt,
rightmargin=0pt}
\AtBeginDocument{%
  \providecommand\BibTeX{{%
    \normalfont B\kern-0.5em{\scshape i\kern-0.25em b}\kern-0.8em\TeX}}}

\setcopyright{acmcopyright}
\copyrightyear{2018}
\acmYear{2018}
\acmDOI{10.1145/1122445.1122456}

\acmConference[Woodstock '18]{Woodstock '18: ACM Symposium on Neural
  Gaze Detection}{June 03--05, 2018}{Woodstock, NY}
\acmBooktitle{Woodstock '18: ACM Symposium on Neural Gaze Detection,
  June 03--05, 2018, Woodstock, NY}
\acmPrice{15.00}
\acmISBN{978-1-4503-XXXX-X/18/06}

\begin{document}

\title{Selective Test-Time Compute Scaling for Click-Through Rate Prediction via Uncertainty-Triggered Feature Path Exploration}

\author{Moyu Zhang}
\affiliation{%
  \institution{Alibaba Group}
  \city{Beijing}
  \country{China}
}
\email{zhangmoyu@butp.cn}

\author{Yun Chen}
\affiliation{%
  \institution{Alibaba Group}
  \city{Beijing}
    \country{China}
}
\email{jinuo.cy@alibaba-inc.com}

\author{Yujun Jin}
\affiliation{%
  \institution{Alibaba Group}
  \city{Beijing}
  \country{China}
}
\email{jinyujun.jyj@alibaba-inc.com}

\author{Jinxin Hu}
\authornote{Corresponding Author}
\affiliation{%
  \institution{Alibaba Group}
  \city{Beijing}
  \country{China}
}
\email{jinxin.hjx@alibaba-inc.com}

\author{Yu Zhang}
\affiliation{%
  \institution{Alibaba Group}
  \city{Beijing}
  \country{China}
}
\email{daoji@alibaba-inc.com}

\author{Xiaoyi Zeng}
\affiliation{%
  \institution{Alibaba Group}
  \city{Beijing}
  \country{China}
}
\email{yuanhan@taobao.com}

\begin{abstract}
Scaling test-time compute has proven highly effective for language models, yet the analogous opportunity in industrial Click-Through Rate (CTR) prediction remains largely unexplored. CTR models suffer from a fundamental asymmetry: while training exposes the model to diverse feature combinations, at inference time many combinations are sparsely observed, causing unreliable predictions for a non-trivial fraction of instances. Existing training-phase solutions such as adaptive gating and feature interaction architectures learn a fixed selection function that is itself subject to the same sparsity, offering no per-instance recourse at deployment. We argue that a principled solution requires deciding how much computation each instance deserves \emph{after} observing it, rather than applying a uniform inference procedure to all. To this end, we propose UTTSI (\textbf{U}ncertainty-\textbf{T}riggered \textbf{T}est-\textbf{T}ime \textbf{S}elective \textbf{I}nference), a training-free model-agnostic framework that scales inference depth proportionally to per-instance uncertainty. UTTSI estimates uncertainty from two complementary signals: a model-internal logit confidence score and a data-level frequency prior, which together distinguish epistemic uncertainty due to sparse feature coverage from aleatoric ambiguity near the decision boundary. Based on this estimate, every instance first undergoes adaptive feature filtering that removes embeddings with insufficient training support, and uncertain instances additionally receive a variable number of stochastic feature-path explorations proportional to their estimated uncertainty, whose predictions are aggregated via consistency-weighted ensembling. Confident instances bypass exploration entirely, keeping the average overhead to approximately $2.8\times$ base model cost while maintaining worst-case latency equal to a single forward pass on parallelized serving infrastructure. Extensive experiments demonstrate that UTTSI yields significant improvements over all baselines.
\end{abstract}



\maketitle

\section{Introduction}
\begin{figure*}[t]
  \centering
  \includegraphics[width=\linewidth]{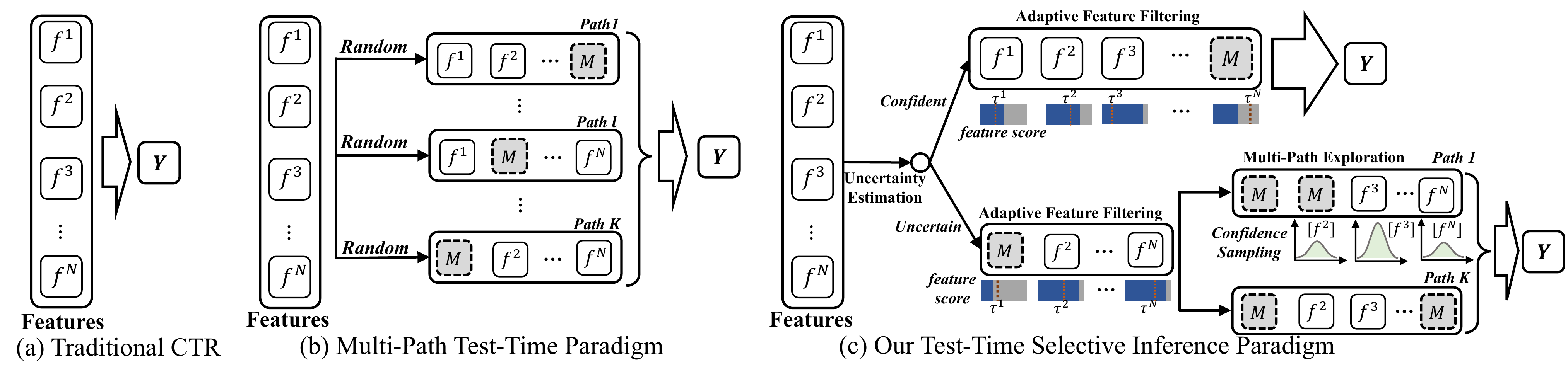}
  \caption{A comparison of CTR inference paradigms. (a) Standard inference feeds all features directly into the model. (b) Naive multi-path exploration applies uniform feature subset sampling to all instances, incurring unnecessary computation for easy ones. (c) UTTSI estimates per-instance uncertainty and routes accordingly: confident instances undergo lightweight feature filtering only, while uncertain instances receive additional multi-path exploration with consistency-weighted aggregation.}
  \label{example}
   \vspace{-0.3cm}
\end{figure*}

Click-through rate (CTR) prediction models are core modules of recommendation systems. They facilitate accurate recommendations by predicting a user's click probability on a target item \cite{back1, back2, back3, back4}, as shown in Figure \ref{example}(a). The prediction is derived from modeling the interactions among a vast combination of input features, encompassing user profiles and item attributes \cite{fm, deepfm, autoint, sfpnet, pepnet}. Therefore, effectively modeling these feature interactions is crucial for accurate prediction in CTR tasks \cite{genctr, dgenctr}. With recent advances in deep learning, numerous studies have proposed increasingly complex model architectures to capture feature correlations \cite{gate1, gate2, hstu}, while others have explored generative paradigms to overcome the limitations of the discriminative binary label space \cite{mtgr, dgenctr, onerec}.

Despite significant advances in model architecture and training objectives \cite{dgenctr, mtgr}, existing CTR research has focused almost exclusively on the training phase, leaving the optimization potential of the inference phase largely unexplored. In practice, even well-trained models exhibit substantially different prediction reliability across instances during inference. For feature combinations that are well-represented in training data, the model produces confident and accurate predictions. However, for combinations that are sparse or previously unseen, the model's learned representations become unreliable, often leading to representation collapse and high prediction uncertainty \cite{emb1, rcola}. This per-instance reliability gap cannot be adequately addressed by training-time mechanisms alone. Although adaptive gating and attention mechanisms \cite{erase, gate1, gate2, pepnet} are designed to dynamically weight features, they learn a single deterministic selection function that is itself subject to the same data sparsity: when the model lacks sufficient training observations for a feature combination, its learned gating decisions for that combination are equally unreliable.

More fundamentally, training and inference have inherently asymmetric objectives. The training phase benefits from exposure to diverse patterns, including sparse ones, to learn generalizable feature representations. In contrast, the inference phase should prioritize leveraging only the feature combinations that the model has already learned to evaluate reliably, thereby avoiding the disruptive influence of unreliable features on prediction accuracy. This asymmetry motivates test-time optimization as a complementary paradigm to training-phase improvements. Rather than relying on a single learned decision function, test-time optimization can explore multiple plausible feature configurations for a given instance and aggregate predictions from these diverse paths, deriving robustness from the diversity of exploration rather than the accuracy of any single gate. Moreover, by deferring the allocation of additional computation until the actual instance is observed, test-time optimization enables selective budget allocation: investing extra test-time computation only in samples that exhibit high uncertainty, while avoiding unnecessary overhead for confident predictions. This selective property is particularly valuable in industrial systems \cite{gen1, gen2, p5, tiger, lcrec, idrec}, where the escalating training costs of ever-larger models make maximizing the return on already-trained parameters an increasingly important goal.

While recent works have begun to explore the test-time paradigm in recommendation \cite{tta}, these efforts have primarily focused on augmenting user behavior sequences for sequential recommendation and fail to address the fundamental problem of feature combination sparsity that plagues CTR prediction. Sequence augmentation techniques such as item dropout or reordering are inherently inapplicable to CTR, which operates on sparse categorical feature fields rather than dense item sequences; perturbing feature values does not carry the same semantic continuity, and the per-combination reliability differential that drives prediction uncertainty in CTR has no counterpart in sequential settings. Moreover, directly transferring test-time scaling techniques from Large Language Models (LLMs) \cite{ttl1, ttl2} is infeasible, as LLM methods rely on token-level generation likelihoods for confidence assessment, a mechanism absent in CTR models with binary classification objectives.

This gap motivates us to rethink what test-time optimization should look like for CTR prediction. The core challenge is not merely generating multiple feature views, but doing so in a way that is \emph{grounded in per-instance reliability}. As illustrated in Figure~\ref{example}(b), even a straightforward multi-path extension to CTR---which generates diverse feature subsets and aggregates their predictions---fails to address the fundamental issue: without knowing which features are unreliable for a given instance, the exploration paths may repeatedly include noise features, and the aggregation cannot distinguish useful diversity from redundant variation. The severity of this problem is compounded by the power-law frequency distribution typical of industrial CTR data: a small fraction of feature values dominates training observations, while the vast majority appear only a handful of times, and prior work has shown that such tail-value embeddings can be statistically indistinguishable from randomly initialized ones \cite{emb1, rcola}. Developing an effective uncertainty-driven test-time paradigm for CTR therefore requires addressing two challenges: (1) how to efficiently estimate per-instance predictive uncertainty without built-in confidence signals, and (2) how to explore beneficial feature configurations within strict industrial latency constraints.

To address these challenges, we propose the \textbf{U}ncertainty-\textbf{T}riggered \textbf{T}est-\textbf{T}ime \textbf{S}elective \textbf{I}nference framework (UTTSI). UTTSI is a training-free model-agnostic framework that operates on top of any already-trained CTR model without modifying its parameters, retraining, or altering the model architecture. At inference time, it selectively determines how much additional computation each instance receives based on its estimated uncertainty, as illustrated in Figure \ref{example}(c). UTTSI comprises three tightly integrated components:

\textbf{Frequency Prior Estimation.} UTTSI maintains a probabilistic hashing structure \cite{cms} to quantify how well-observed each feature value is in training data, providing an offline data-level reliability signal to the subsequent uncertainty estimation step.

\textbf{Dual-Signal Uncertainty Estimation.} UTTSI combines an importance-weighted frequency prior with model-internal logit confidence to produce a per-instance uncertainty score, separating epistemic uncertainty (sparse feature coverage) from aleatoric ambiguity (inherent label noise), and continuously allocates exploration paths proportional to the estimated uncertainty.

\textbf{Feature Filtering and Path Exploration.} Every instance first undergoes adaptive feature filtering, removing features with unreliable representations via per-field thresholds computed offline. For uncertain instances, UTTSI generates additional inference paths by stochastically sampling diverse subsets of the refined features, with sampling probabilities guided by a composite score of frequency-based reliability and attribution strength. Predictions from all paths are aggregated via a consistency-weighted scheme; confident instances bypass multi-path exploration entirely.

The contributions of our paper can be summarized as follows:
\begin{itemize}
\item As far as we know, UTTSI is the first training-free model-agnostic test-time framework, which improves performance by selectively scaling inference depth proportional to per-instance uncertainty.
\item UTTSI introduces a dual-signal uncertainty estimator combining logit confidence with a frequency-based prior, an adaptive feature filtering mechanism that removes unreliable representations for all instances, and a path-allocation mechanism that derives robustness from feature path exploration for uncertain ones.
\item Experiments across four datasets with three backbone architectures and a seven-day online A/B test demonstrate that UTTSI achieves consistent, statistically significant improvements over state-of-the-art CTR models, confirming its broad compatibility and practical value.
\end{itemize}

\section{Related Work}
\subsection{CTR Prediction Models}
Deep learning has driven substantial advances in CTR prediction, with a large body of work focused on designing sophisticated feature interaction architectures \cite{rel1}. Representative models include WDL \cite{wdl}, DeepFM \cite{deepfm}, DCN \cite{dcn, dcn2}, and xDeepFM \cite{xdeepfm}, which capture low- and high-order feature crosses through various network designs. Subsequent work introduced gating mechanisms for dynamic feature selection \cite{gate1, gate2} and embedding calibration \cite{erase, pepnet}, while HSTU \cite{hstu} scales to large-scale user action sequences. Recognizing limitations of the binary discriminative objective, a separate line of work reframes CTR via generative paradigms \cite{mtgr, dgenctr}. However, all these methods focus exclusively on training-phase optimization, overlooking the potential of the test-time computation.

\subsection{Test-Time Optimization}
Test-time scaling has driven significant gains in LLMs \cite{ttl1, ttl2, ttl3, ttl4} through iterative refinement and hypothesis search, establishing a favorable trade-off between test-time compute and prediction quality. However, these methods are tightly coupled to autoregressive architectures and token-level confidence signals, making them non-transferable to CTR models with binary classification objectives. In the recommendation domain, TTA \cite{tta} augments user behavior sequences at test time for sequential recommendation, but sequence augmentation is inapplicable to CTR: CTR operates on sparse categorical feature fields where per-combination reliability differentials drive prediction uncertainty, a problem absent in sequential settings. More broadly, adaptive computation methods such as early-exit networks \cite{adaptcomp1} allocate varying computation by modifying model depth or width, whereas UTTSI operates on the input feature space, making it architecture-agnostic and applicable to any already-trained CTR model without retraining. In summary, UTTSI is the first test-time optimization framework specifically designed for CTR prediction, addressing feature combination sparsity and classification uncertainty through feature path exploration.

\section{Preliminary}
This section defines the CTR prediction task and introduces the problem setting for test-time optimization in CTR. UTTSI operates entirely at inference time on an already-trained model, requiring no parameter updates or additional training data.

\subsection{CTR Prediction Task}\label{sec:ctr}

The Click-Through Rate (CTR) prediction task estimates the probability that a user will click on a presented item, and its output determines the final ranking of items shown to users. It is formulated as a supervised binary classification problem.

\textbf{Problem Definition.}  Given a complete set of features $\textbf{F}_{full}$ and the label space $y \in \{0, 1\}$, the CTR prediction task learns a ranking function $\mathcal{F}: \textbf{F}_{full} \rightarrow y$ to estimate the click probability of the target user on the target item. The feature space encompasses various fields such as user profiles and item attributes. We define $\textbf{F}_{full} = [f^1, f^2, \ldots, f^N]$, where $N$ is the number of feature fields:
\begin{gather}
P(y| \textbf{F}_{full}) =  \mathcal{F}(f^1, f^2, ..., f^{N})
\end{gather} 

\begin{figure*}[t]
  \centering
  \includegraphics[width=\linewidth]{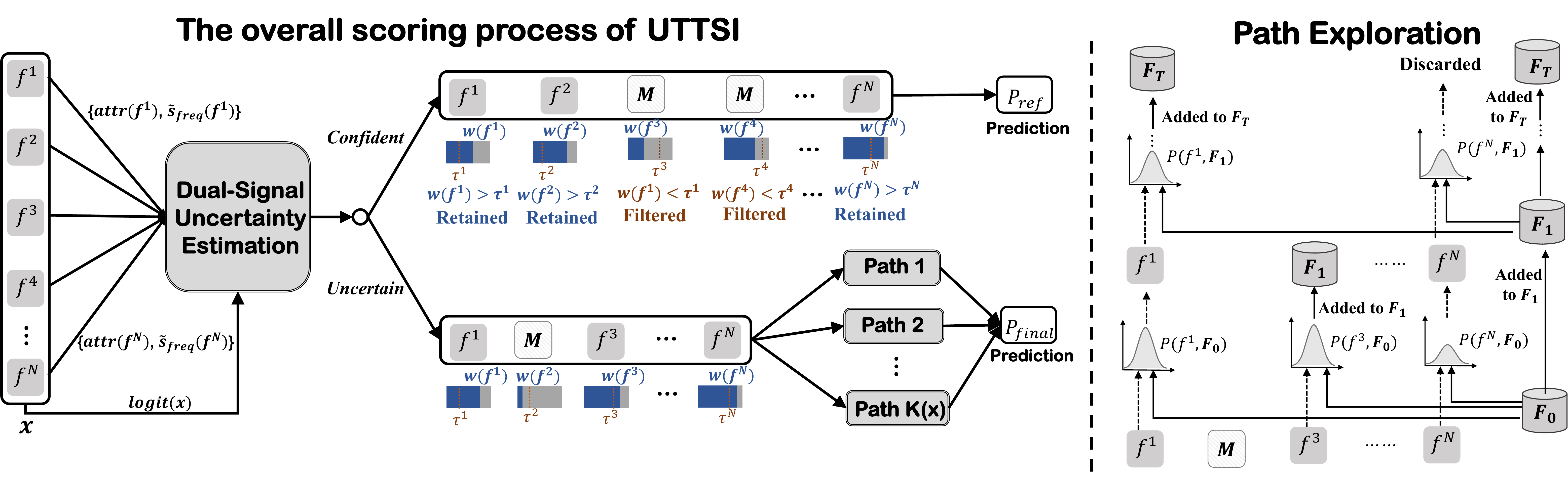}
  \caption{The structure of Uncertainty-Triggered Test-Time Selective Inference framework (UTTSI) for CTR models. The left part shows the overall UTTSI scoring process including uncertainty-based routing, and the right part shows the specific inference path generation process for uncertain instances.}
  \label{model}
   \vspace{-0.3cm}
\end{figure*}

\subsection{Test-Time Optimization in CTR}
\textbf{Problem Setting.} We consider a CTR model $\mathcal{F}_{\theta}$ whose parameters are fixed after training, as defined in Section~\ref{sec:ctr}. Test-time optimization augments inference under three constraints: no updates to $\theta$, no new learnable parameters, and no access to training labels.

\textbf{Test-Time Optimization Paradigm.} A test-time optimization method augments standard inference by wrapping $\mathcal{F}_\theta$ with an additional procedure. In the multi-path setting, the method generates a collection of $K$ feature views $[\textbf{F}^1, \textbf{F}^2, \ldots, \textbf{F}^K]$ derived from $\textbf{F}_{full}$, obtains a prediction from each view, and aggregates them into a final prediction score during inference time:
\begin{gather}
\hat{y}_{\text{TTO}} = \textsc{Agg}\!\left(\left\{\mathcal{F}_\theta(\textbf{F}^k)\right\}_{k=1}^{K}\right),
\end{gather}
where $\textsc{Agg}(\cdot)$ is an aggregation function such as mean or consistency-weighted average, and $K$ may be fixed or instance-adaptive. The key design choices are: (1) how to generate diverse yet semantically coherent feature views $\textbf{F}^k$; (2) how to determine the number of paths $K$ per instance; and (3) how to weight the predictions in aggregation. A naive approach applies uniform random feature dropout for all instances with a fixed $K$, as illustrated in Figure~\ref{example}(b). This ignores instance-level reliability differences and wastes computation on instances whose base prediction is already reliable.

An uncertainty-aware instantiation addresses this by making all three choices contingent on a per-instance reliability signal $u(x) \in [0,1]$ estimated from the frozen model. It first applies a lightweight feature filtering pass to all instances to remove unreliable representations, then allocates additional exploration paths proportional to $u(x)$, and finally weights path predictions by their mutual consistency. UTTSI, described in Section~\ref{sec:method}, provides principled solutions for each of these steps.

\section{Method}\label{sec:method}
We propose a training-free Uncertainty-Triggered Test-Time Selective Inference framework (UTTSI) that operates on top of a frozen, already-trained CTR model $\mathcal{F}_\theta$ without modifying its parameters. After estimating per-instance uncertainty via dual signals, UTTSI applies adaptive feature filtering to all instances and selectively scales inference depth proportional to each instance's uncertainty, so that uncertain instances receive deeper exploration while confident ones incur minimal overhead. Specifically, UTTSI comprises three tightly integrated components:\\
\textbf{$\bullet$ Frequency Prior Estimation}: Estimates how well-observed each feature value in the training dataset is using a Count-Min Sketch inspired structure \cite{cms}, providing an offline data-level reliability signal for uncertainty estimation.\\
\textbf{$\bullet$ Dual-Signal Uncertainty Estimation}: Combines an attribution-weighted frequency prior with model logit confidence to produce a per-instance uncertainty score, which continuously determines how many exploration paths each instance receives.\\
\textbf{$\bullet$ Feature Filtering and Path Exploration}: Applies adaptive feature filtering to all instances, removing features below a per-field reliability threshold. For uncertain instances, additional stochastic exploration paths are generated over the refined feature subset and aggregated via consistency-weighted voting; confident instances output the filtered prediction directly.

\subsection{Frequency Prior Estimation}\label{sec:freq}
A key component of our uncertainty estimation is the data-level frequency prior, which quantifies how well-observed each feature value is in the training data. The rationale is straightforward: a feature value that appears frequently during training yields a well-trained embedding, whereas a rare value produces an unreliable one. We emphasize that frequency serves as one signal within our dual-signal framework, not as a standalone confidence measure.

Since the number of unique feature values can be very large, we adopt a Count-Min Sketch \cite{cms} inspired probabilistic hashing structure that maintains $L$ independent hash tables per feature field, estimating the count of each value by taking the minimum across tables, which provides a tight upper bound while mitigating overestimation from hash collisions. The resulting per-field frequency counts $\text{cnt}(f^i)$ are precomputed offline and retrieved at negligible online cost. We normalize each count by a saturation threshold $\eta$ to obtain $\tilde{s}_{freq}(f^i) = \min(\text{cnt}(f^i), \eta) / \eta \in [0,1]$, which is used in both the uncertainty estimation (\S\ref{sec:dual}) and the feature filtering (\S\ref{sec:refine}).

\subsection{Dual-Signal Uncertainty Estimation}\label{sec:dual}
This module determines the amount of selective inference computation allocated to each instance. The key insight is that neither model confidence nor data frequency alone reliably identifies uncertain instances. A prediction logit close to zero may reflect genuine epistemic uncertainty (the model lacks knowledge) or aleatoric ambiguity (the true click probability is near 50\%). Frequency alone flags instances with rare feature values, but cannot distinguish reliable from unreliable predictions on unseen combinations where individual features are frequent yet their interaction pattern is novel. By combining both signals, we can better identify instances that genuinely benefit from additional exploration.

For a given input instance $x$ with feature set $\textbf{F}_{full}$, we first perform a standard forward pass and compute the prediction logit:
\begin{gather}
\text{logit}(x) = \mathcal{F}_\theta(\textbf{F}_{full})
\end{gather}
We then backpropagate from the output logit to obtain the gradient norm for each feature embedding:
\begin{gather}
\text{attr}(f^i) = \|\nabla_{e_i} \text{logit}(x)\|_2
\end{gather}
This computation is fully model-agnostic: it requires only that the backbone is differentiable, and no model parameters are updated. The backward pass computes only the input embedding gradients (not model parameter gradients), which incurs approximately $0.5$--$1\times$ the cost of a forward pass depending on model architecture. The gradient norm measures how much a small perturbation to the feature embedding would change the prediction, providing a principled, unified measure of feature influence across arbitrary architectures. This attribution is reused in both the uncertainty estimation below and the feature filtering in \S\ref{sec:refine}.

The \textbf{model-internal confidence} captures how decisive the model's prediction is:
\begin{gather}
s_{model}(x) = \min\left(\frac{|\text{logit}(x)|}{\gamma}, \, 1\right)
\end{gather}
where $\gamma$ is a normalization constant set to the 95th percentile of $|\text{logit}|$ values on a held-out validation set. A value near 0 indicates the model is at its decision boundary; a value near 1 indicates a decisive prediction.

The \textbf{data-level frequency confidence} is an attribution-weighted average of the per-field frequencies:
\begin{gather}
s_{freq}(x) = \frac{\sum_{i=1}^{N} \text{attr}(f^i) \cdot \tilde{s}_{freq}(f^i)}{\sum_{i=1}^{N} \text{attr}(f^i)}
\end{gather}
where $\tilde{s}_{freq}(f^i)$ is the normalized frequency of feature value $f^i$ from the offline index (defined in \S\ref{sec:freq}). By weighting each feature's frequency by its attribution, $s_{freq}$ reflects not merely how frequent the sample's features are on average, but how frequent the features that actually drive the prediction are. This attribution weighting addresses a key blind spot: if an instance contains high-frequency but low-attribution features alongside low-frequency but high-attribution features, an unweighted average would misleadingly inflate confidence, whereas the weighted sum correctly identifies the sample as uncertain because the predictive signal comes from poorly trained embeddings.

The per-instance uncertainty is their complement:
\begin{gather}
u(x) = 1 - \left[\alpha \cdot s_{model}(x) + (1-\alpha) \cdot s_{freq}(x)\right]
\end{gather}
where $\alpha \in [0,1]$ balances the two signals. The number of additional inference paths is then:
\begin{gather}
K(x) = \lfloor K_{\max} \cdot u(x) \rfloor
\end{gather}
where $K_{\max}$ is a hyperparameter that sets the maximum number of exploration paths any instance can receive, directly controlling the worst-case computational overhead.

\textbf{Intuition.} The continuous allocation $K(x) = \lfloor K_{\max} \cdot u(x) \rfloor$ has a desirable property: instances with higher estimated uncertainty receive more exploration paths. Since $u(x)$ is designed to correlate positively with prediction variance, allocating more paths to uncertain instances preferentially reduces variance where it is largest. The computational overhead is bounded by $1 + K_{\max} \cdot \mathbb{E}[u(x)]$, which empirically amounts to approximately $2$-$3\times$ base model cost.

\subsection{Feature Filtering and Path Exploration}\label{sec:refine}
This module implements the core inference mechanism of UTTSI in two stages. A natural question is why uncertainty estimation precedes filtering rather than the reverse, given that filtering applies uniformly to all instances. There are two reasons. First, the gradient attribution $\text{attr}(f^i)$ computed during uncertainty estimation is reused by the filtering step to form the composite score $w(f^i)$; reversing the order would require a second backward pass on the filtered input, adding unnecessary computation. Second, and more fundamentally, uncertainty should reflect the instance's raw difficulty including the presence of unreliable features. An instance with many low-frequency features is inherently uncertain, and filtering those features first would artificially deflate the uncertainty score, potentially under-allocating exploration budget for instances that still suffer from combination-level uncertainty after noise removal.

\textbf{Stage 1 -- Adaptive Feature Filtering.} Using the gradient attribution $\text{attr}(f^i) = \|\nabla_{e_i} \text{logit}(x)\|_2$ computed above, we form the composite score $w(f^i) = \beta \cdot \tilde{s}_{freq}(f^i) + (1-\beta) \cdot \text{attr}(f^i)$ for each feature. This composite score leverages both the data-level frequency signal (how well-observed the feature value is) and the model-level attribution signal (how influential the feature is for the current prediction), ensuring that features identified as noise on both dimensions are candidates for removal.

Rather than applying a fixed per-instance retention ratio or a single global threshold, we adopt a \emph{per-field threshold} $\tau^i$ for each feature field $i$, computed offline from the training data. This is motivated by the strong heterogeneity of features in industrial recommendation systems: the composite score distributions of user identity fields, item attribute fields, and statistical context fields differ substantially, and a single global threshold would systematically over-filter high-variance fields while under-filtering low-variance ones. Specifically, we compute $w(f^i)$ for all training instances within each field separately and set $\tau^i$ to the $\rho$-th quantile of that field's score distribution, where $\rho \in (0,1)$ is a shared hyperparameter controlling the aggressiveness of filtering uniformly across fields. At inference time, a feature $f^i$ is retained if $w(f^i) \geq \tau^i$; an instance where all features exceed their respective thresholds is preserved in full. The retained features form a refined feature set $\hat{\textbf{F}}$, and a forward pass on $\hat{\textbf{F}}$ produces a refined prediction:
\begin{gather}
P_{ref} = \mathcal{F}_\theta(\hat{\textbf{F}})
\end{gather}

This filtering strategy is \emph{conservative}: a feature is discarded only when it scores low on \emph{both} frequency and attribution. Features with conflicting signals (e.g., low frequency but high attribution) are retained, since the cost of discarding a genuinely informative rare feature outweighs the cost of retaining noise, which Stage~2's multi-path aggregation can average out.

\textbf{Why filtering alone suffices for high-confidence instances.} A natural concern is that instances with $K(x)=0$ receive only filtering without multi-path compensation, which might appear to lose information. We argue that this asymmetric design is justified because filtering and exploration address distinct error modes at different levels. Feature filtering removes \emph{feature-level noise} (individual features whose embeddings are unreliable), which benefits all instances uniformly. However, after filtering, low-confidence instances still suffer from \emph{combination-level uncertainty}: even though each retained feature is individually reliable, their joint interaction pattern remains sparsely observed in training, so the model's prediction on $\hat{\textbf{F}}$ has high variance. Multi-path exploration averages this \emph{residual variance} by sampling diverse subsets of $\hat{\textbf{F}}$ and aggregating their predictions. High-confidence instances, by contrast, have low residual variance: the model has already learned reliable representations for their feature combinations, so a single refined prediction suffices. This bias-variance decomposition justifies the asymmetric design: filtering reduces bias for all instances, while the multi-path feature exploration reduces residual variance only where it remains significant.

\textbf{Stage 2 -- Feature Path Exploration (for $K(x)>0$).} For instances with non-negligible uncertainty ($K(x)>0$), UTTSI generates $K(x)$ additional inference paths by stochastically perturbing the refined feature set $\hat{\textbf{F}}$. Unlike the original feature set which may include unreliable feature combinations that compromise prediction accuracy \cite{emb1}, our goal is to construct multiple high-quality feature subsets and aggregate their predictions for robustness.

Because there is no absolute threshold distinguishing reliable from unreliable feature combinations, we adopt stochastic sampling across multiple paths rather than deterministic selection, ensuring diverse exploration that preserves potentially effective feature information. A path is built iteratively over $T$ steps: at each step $t$, the algorithm samples from the features not yet included in the current path $\boldsymbol{F}_{t-1}$, with each candidate's probability proportional to its composite reliability--attribution score:
\begin{gather}
p_i^t = \frac{w(f^i)}{\sum_{f^j \in \hat{\textbf{F}},\, f^j \not\in \boldsymbol{F}_{t-1}} w(f^j)}
\end{gather}
where $w(f^i) = \beta \cdot \tilde{s}_{freq}(f^i) + (1-\beta) \cdot \text{attr}(f^i)$, with $\tilde{s}_{freq}(f^i)$ the normalized per-field frequency and $\text{attr}(f^i) = \|\nabla_{e_i} \text{logit}(x)\|_2$ the gradient norm from Stage~1. The hyperparameter $\beta$ balances the two components. This ensures that informative rare features with strong attribution are preferentially selected, while features that are both infrequent and weakly attributed are more likely to be excluded from any given path.

For each candidate feature, we perform an independent Bernoulli sampling using its corresponding normalized sampling probability to decide whether to add it to the path:
\begin{gather}
\mathbb{I}^t_i = 
\begin{cases}
f^i, \,  r^t_i= 1 \\
\varnothing, \, r^t_i = 0
\end{cases}  \\
\boldsymbol{F}_t = \boldsymbol{F}_{t-1} \cup \left\{ \mathbb{I}^t_i \right\}_{f^i \in \hat{\textbf{F}}}^{f^i \not\in \boldsymbol{F}_{t-1}}
\end{gather} 
where $r_i^t \sim Bernoulli(p_i^t)$ denotes the Bernoulli sampling on $f^i$.

Upon completion of $T$ iterations, this procedure yields an instance-specific feature set $\boldsymbol{F}_T$ derived from the refined base $\hat{\textbf{F}}$. A new input is constructed by masking all features not selected for this path, which is then passed to the trained CTR prediction model:
\begin{gather}
\mathcal{G}(f^i) = 
\begin{cases}
f^i, \,  f^i \in \boldsymbol{F}_T \\
\textbf{0}, \, f^i \not \in \boldsymbol{F}_T
\end{cases}  \\
P(y|\boldsymbol{F}_T) = \mathcal{F}_\theta(\left\{ \mathcal{G}(f^i) \right\}_{f^i \in \hat{\textbf{F}}} )
\end{gather} 

Since a single stochastic path is not guaranteed to find the globally optimal feature combination, we sample $K(x)$ independent paths in parallel, producing a set of distinct final feature sets $L = \left\{ \boldsymbol{F}_T^1, \boldsymbol{F}_T^2, ..., \boldsymbol{F}_T^{K(x)}\right\}$. Combined with the refined prediction $P_{ref}$, we obtain $K(x)+1$ prediction scores in total. This multi-path approach allows for a more comprehensive exploration of the feature space, with the inference breadth automatically scaling with the instance's uncertainty via the selective allocation $K(x)$.

\textbf{Consistency-Weighted Aggregation.} Given predictions from the $K(x)$ exploration paths plus the refined prediction $P_{ref} = \mathcal{F}_\theta(\hat{\textbf{F}})$, we aggregate all $K(x)+1$ scores via consistency-weighted voting. Let $\mathcal{P} = \{P_{ref}\} \cup \{P(y|\boldsymbol{F}_T^k)\}_{k=1}^{K(x)}$ denote the full set of predictions. We then weight each prediction of multi-paths by its consistency with the ensemble mean:
\begin{gather}
\bar{P} = \frac{1}{K(x)+1}\sum_{P_i \in \mathcal{P}} P_i \\
w_i = \exp\left(-\lambda \cdot |P_i - \bar{P}|\right) \\
P_{final} = \frac{\sum_{P_i \in \mathcal{P}} w_i \cdot P_i}{\sum_{P_i \in \mathcal{P}} w_i}
\end{gather}
where $\bar{P}$ is the mean prediction across all paths including the base, $\lambda$ controls the sharpness of the consistency weighting, and $w_i$ assigns higher weight to predictions aligned with the majority. Paths that produce outlier predictions, likely due to having masked critical features, are naturally down-weighted. When $K(x) = 0$, no additional paths are generated and $P_{final} = P_{ref}$ (the refined prediction from Stage 1). This ensures that the final output reflects a robust consensus across diverse selective inference paths, with the depth of inference, measured by the number of ensembled predictions, automatically scaled to the instance's uncertainty: no exploration for confident instances ($K(x)=0$), and progressively deeper exploration for uncertain ones ($K(x)>0$).

\textbf{Sensitivity to $\lambda$.} The sharpness parameter $\lambda$ controls how aggressively outlier paths are down-weighted; detailed sensitivity analysis is provided in Section~\ref{sec:rq3}. We set $\lambda = 5$ as the default.

\begin{algorithm}[t]
\small
\caption{UTTSI Inference Procedure}
\label{alg:utte}
\KwIn{Trained CTR model $\mathcal{F}_\theta$; input instance with feature set $\textbf{F}_{full}$; precomputed per-field frequency counts $\text{cnt}(\cdot)$; per-field thresholds $\{\tau^i\}$ (precomputed offline); hyperparameters $K_{\max}$, $\alpha$, $\beta$, $T$, $\gamma$, $\eta$, $\lambda$}
\KwOut{Final CTR prediction $P_{final}$}

\tcp{Step 1: Base forward pass + attribution}
$\text{logit}(x) \leftarrow \mathcal{F}_\theta(\textbf{F}_{full})$\;
Compute $\text{attr}(f^i) = \|\nabla_{e_i} \text{logit}(x)\|_2$ for all $f^i \in \textbf{F}_{full}$ via backpropagation\;

\tcp{Step 2: Dual-signal uncertainty estimation}
$s_{model}(x) \leftarrow \min\!\left(|\text{logit}(x)| / \gamma,\; 1\right)$\;
$s_{freq}(x) \leftarrow \frac{\sum_{i=1}^{N} \text{attr}(f^i) \cdot \tilde{s}_{freq}(f^i)}{\sum_{i=1}^{N} \text{attr}(f^i)}$\;
$u(x) \leftarrow 1 - \left[\alpha \cdot s_{model}(x) + (1-\alpha) \cdot s_{freq}(x)\right]$\;
$K(x) \leftarrow \lfloor K_{\max} \cdot u(x) \rfloor$\;

\tcp{Step 3: Adaptive feature filtering}
Compute composite weights $w(f^i) = \beta \cdot \tilde{s}_{freq}(f^i) + (1-\beta) \cdot \text{attr}(f^i)$ for all $f^i$\;
Retain features with $w(f^i) \geq \tau^i$ to form $\hat{\textbf{F}}$\;
$P_{ref} \leftarrow \mathcal{F}_\theta(\hat{\textbf{F}})$\;

\If{$K(x) = 0$}{
  \Return $P_{ref}$\;
}

\tcp{Step 4: Feature path exploration (parallelizable)}
$\mathcal{P} \leftarrow \{P_{ref}\}$\;
\For{$k \leftarrow 1$ \KwTo $K(x)$ \textbf{(in parallel)}}{
  $\boldsymbol{F}_0^k \leftarrow \emptyset$\;
  \For{$t \leftarrow 1$ \KwTo $T$}{
    Compute path weights $w(f^i)$ for unselected $f^i \in \hat{\textbf{F}}$\;
    Sample $r_i^t \sim \text{Bernoulli}(p_i^t)$ for each unselected feature\;
    $\boldsymbol{F}_t^k \leftarrow \boldsymbol{F}_{t-1}^k \cup \{f^i : r_i^t = 1\}$\;
  }
  Construct masked input $\{\mathcal{G}(f^i)\}_{f^i \in \hat{\textbf{F}}}$ using $\boldsymbol{F}_T^k$\;
  $P^k \leftarrow \mathcal{F}_\theta(\{\mathcal{G}(f^i)\})$\;
  $\mathcal{P} \leftarrow \mathcal{P} \cup \{P^k\}$\;
}

\tcp{Step 5: Consistency-weighted aggregation}
$\bar{P} \leftarrow \frac{1}{|\mathcal{P}|}\sum_{P_i \in \mathcal{P}} P_i$\;
$w_i \leftarrow \exp(-\lambda \cdot |P_i - \bar{P}|)$ for each $P_i \in \mathcal{P}$\;
$P_{final} \leftarrow \sum_{P_i \in \mathcal{P}} w_i P_i \;/\; \sum_{P_i \in \mathcal{P}} w_i$\;
\Return $P_{final}$\;
\end{algorithm}

\section{Experiments}
We evaluate our UTTSI on several benchmark datasets to address the following research questions:\\
\textbf{$\bullet$ RQ1:} Does UTTSI consistently improve prediction performance when applied to state-of-the-art CTR models? \\
\textbf{$\bullet$ RQ2:} What is the contribution of each component within UTTSI? \\
\textbf{$\bullet$ RQ3:} How sensitive is UTTSI to its key hyperparameters, and how does uncertainty-triggered allocation behave in practice? \\
\textbf{$\bullet$ RQ4:} Does the uncertainty score correlate with prediction error, and how does UTTSI perform across different uncertainty strata?

\subsection{Datasets}
We evaluate on four large-scale datasets: three public benchmarks and one industrial dataset. Statistics are in Table~\ref{datasets}.\\
\textbf{$\bullet$ Criteo  \footnote{http://labs.criteo.com/downloads/download-terabyte-click-logs/}}.  It is a canonical public benchmark for evaluating CTR prediction models \cite{criteo}. It consists of one week of real-world ad click data, featuring 13 continuous features and 26 categorical features.\\ 
\textbf{$\bullet$ Avazu  \footnote{http://www.kaggle.com/c/avazu-ctr-prediction}}. It is another widely-used public benchmark for CTR prediction, consisting of 10 days of chronologically ordered ad click logs \cite{avazu}. It features 23 feature fields in a sample. \\
\textbf{$\bullet$ KDD12  \footnote{http://www.kddcup2012.org/c/kddcup2012-track2/data}}. This dataset contains training instances derived from search session logs. It has 11 categorical fields, and the click field is the number of times the user clicks the ad. \\
\textbf{$\bullet$ Industrial Dataset}. To assess our approach in a real-world setting, we gathered a dataset from an international e-commerce platform's online display advertising system. The training set is composed of samples from the last 20 days, while the test set consists of exposure samples from the subsequent day.

\begin{table}[t]
	\small
	\centering
	\begin{tabular}{cccc}
    		\hline Dataset &  \# Feature Fields & \# Impressions & \# Positive \\
		\hline Criteo & 39 & 45M & 26\%  \\
		Avazu& 23 & 40M & 17\% \\
		KDD12 & 11 & 60M& 4.5\% \\
		Industrial. & 68 & 513M & 2.5\% \\
		\hline
	\end{tabular}
	\caption{Statistics of four benchmark datasets.}
	\label{datasets}
\end{table} 

\begin{table*}[t]
	\caption{Prediction performance of CTR models on four datasets. $\Delta_{AUC}$ and $\Delta_{Logloss}$ indicate the average performance change relative to DeepFM. * indicates p-value $<$ 0.05 in the significance test.}
	\begin{tabular}{c|cc|cc|cc|cc|cc}
    \toprule
      \multirow{2}{*}{\diagbox{Method}{Dataset}}& \multicolumn{2}{c|}{Criteo} & \multicolumn{2}{c|}{Avazu} &\multicolumn{2}{c|}{KDD12}& \multicolumn{2}{c|}{Industrial.}&$\Delta_{AUC}$&$\Delta_{Logloss} $\cr
   \cmidrule(lr){2-9}
    & AUC& Logloss& AUC& Logloss& AUC& Logloss& AUC&{Logloss}& $\uparrow$ &$\downarrow$\cr 
    \midrule
    FM &0.7695  & 0.4716  & 0.7758  &0.4469  &0.7914  &0.1431 &0.7788  &0.0852 &-0.03\%&+0.0003   \cr
    DNN &0.7657  & 0.4739  & 0.7699  &0.4475  &0.7902  &0.1437 &0.7801  &0.0849 &-0.34\%&+0.0011  \cr
    Wide\&Deep &0.7677  & 0.4728  & 0.7755  &0.4465  &0.7921  &0.1426 &0.7783  &0.0854 &-0.09\%&+0.0004  \cr
	DeepFM &0.7692  &0.4713 &0.7756 &0.4469  &0.7933  & 0.1422  &0.7785  & 0.0852 & -  & -   \cr
	DCN &0.7703  &0.4703 &0.7762 &0.4458  &0.7941  & 0.1426  &0.7792  & 0.0851 & +0.11\%&-0.0005   \cr
 AutoInt & 0.7695  & 0.4710 &0.7748  &0.4473  &0.7928  &0.1429  &0.7823  &0.0847  &+0.09\%&-0.0001   \cr
	FibiNET  &0.7732  &0.4691  &0.7759 &0.4456  &0.7968   &0.1402  &0.7825  &0.0844  & +0.38\% &-0.0016  \cr
	GDCN &0.7796  &0.4663  &0.7809  &0.4428  &0.7989  &0.1399 &0.7884  &0.0839  &+1.00\%&-0.0032  \cr
	MaskNet &0.7882 &0.4644 &0.7813  &0.4415  &0.8012  &0.1381 &0.7846  &0.0831  &+1.25\% &-0.0046  \cr
PEPNet &0.7981 &0.4498  &0.7944  &0.4402  &0.8041  &0.1370 &0.7904  &0.0817 &+2.27\% &-0.0092  \cr
	HSTU &0.7993&0.4483  &0.7902  &0.4403  &0.8087  &0.1358  &0.7926  &0.0814  &+2.39\% &-0.0099 \cr
	\midrule
	 MMOE &0.7906  & 0.4565  &0.7879 &0.4420  &0.8021  &0.1377  &0.7899  &0.0826  &+1.74\% &-0.0067  \cr
	PLE &0.7937  &0.4536  &0.7856  &0.4441  &0.8035  &0.1370  &0.7905  &0.0823  &+1.83\%&-0.0072 \cr
	\midrule
	AutoCTR &0.7987  &0.4487 &0.7908  &0.4396  &0.8029  &0.1373  &0.7893  &0.0828  &+2.10\% &-0.0093 \cr
	OptFu &0.8014  &0.4480  &0.7924  &0.4383 &0.8125  &0.1326  &0.7933  &0.0803  &+2.66\% &-0.0116 \cr
	\midrule
	 HSTU+UTTSI &0.8041& 0.4452 & 0.7963 & 0.4351 & 0.8137 & 0.1314 &0.7969 &0.0790 &+3.04\% &-0.0137  \cr
	 PLE+UTTSI &0.7998 & 0.4473 & 0.7916 & 0.4390 & 0.8097 &  0.1338  &0.7954 &0.0796 &+2.57\% &-0.0115 \cr
	 \textbf{OptFu+UTTSI}&\textbf{0.8051*} & \textbf{0.4448*} & \textbf{0.7972*} & \textbf{0.4359*} &  \textbf{0.8169*} &  \textbf{0.1297*} &\textbf{0.7979*} &\textbf{0.0786*} &\textbf{+3.23}\% &\textbf{-0.0142}  \cr
    \bottomrule
    \end{tabular}
	\label{results}
	\vspace{-0.1cm}
\end{table*} 

\subsection{Competitors}
To demonstrate the effectiveness of UTTSI, we compare against state-of-the-art CTR models grouped into three categories: 1) \textbf{Deep Feature Interaction Models}: FM \cite{fm}, DNN, Wide\&Deep \cite{wdl}, DeepFM \cite{deepfm}, DCN \cite{dcn}, AutoInt \cite{autoint}, FiBiNet \cite{fibinet}, GDCN \cite{gdcn}, MaskNet \cite{masknet}, PEPNet \cite{pepnet}, and HSTU \cite{hstu}. 2) \textbf{Multi-Expert Architectures}: MMoE \cite{mtl3} and PLE \cite{ple}. 3) \textbf{Neural Architecture Search Models}: AutoCTR \cite{autoctr} and OptFu \cite{dag}.

\textbf{Implementation Details}.  All models were implemented in TensorFlow \cite{tensorflow} and trained on 8 NVIDIA A100 GPUs using the Adam optimizer \cite{adam} and Xavier initialization \cite{xavier}. The default activation function was ReLU. Optimal hyperparameters were identified via grid search. Embedding dimensions were set to 32 (Criteo, Avazu), 16 (KDD12), and 8 (Industrial), with a batch size of 4096. Learning rates were searched in $\{3\text{e-}3, \ldots, 1\text{e-}5\}$ and $L_2$ regularization in $\{3\text{e-}6, \ldots, 0\}$. For UTTSI, we set $K_{\max}=8$ as the default maximum path budget, $\alpha=0.05$ for the dual-signal weighting between logit confidence and frequency prior, $\beta=0.4$ for the composite weight balancing frequency reliability and attribution strength, $\rho=0.2$ as the shared quantile for computing the per-field adaptive filtering thresholds $\{\tau^i\}$, the number of iterative sampling steps $T=10$, and the consistency-weighting sharpness $\lambda=5$. 

\textbf{Evaluation Metrics}. Following standard practice in CTR research, we adopt AUC (Area Under ROC Curve) and LogLoss (binary cross-entropy) as evaluation metrics.

\subsection{Comparison with Baselines (RQ1)}
Table \ref{results} presents the overall prediction performance across all datasets, with statistical significance tests against the best baseline. UTTSI consistently outperforms all baselines on datasets when applied to three representative backbone architectures, achieving this without any modification to the backbone's parameters or retraining.

Several observations are noteworthy. Models with sophisticated cross-feature interaction architectures such as GDCN \cite{gdcn} and MaskNet \cite{masknet} improve over simpler baselines, but their gains plateau as model complexity increases. Multi-expert architectures (MMoE \cite{mtl3}, PLE \cite{ple}) and NAS-based methods (AutoCTR \cite{autoctr}, OptFu \cite{dag}) further improve by diversifying the feature interaction space, yet they share a common limitation: all optimizations are confined to the training phase and applied uniformly to every instance. UTTSI addresses this by selectively allocating additional inference compute per instance: uncertain samples receive multi-path exploration, while confident ones incur no overhead, complementing rather than competing with any training-phase backbone.

The gains are quantitatively consistent and statistically significant ($p < 0.05$). OptFu+UTTSI achieves the best overall results, with AUC improvements of $+0.0037$, $+0.0048$, $+0.0044$, and $+0.0046$ over the strongest baseline OptFu on Criteo, Avazu, KDD12, and Industrial respectively, along with corresponding Logloss reductions. Notably, gains are most pronounced on KDD12 and Industrial, the two datasets with higher feature sparsity and a larger proportion of tail-value features, directly reflecting UTTSI's targeted improvement of sparse-feature instances. In terms of average relative improvement over DeepFM, OptFu+UTTSI achieves $\Delta_{AUC}=+3.23\%$ and $\Delta_{Logloss}=-0.0142$, surpassing all training-phase baselines across all datasets.

\begin{figure*}[t]
  \centering
  \includegraphics[width=\linewidth]{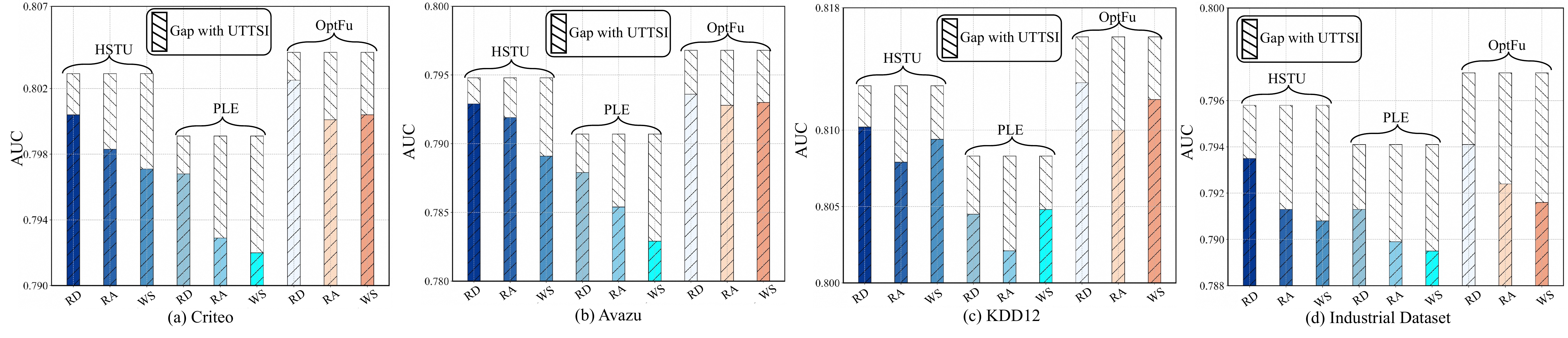}
  \caption{Prediction performance of three variants of UTTSI under three backbones on four datasets.}
  \label{abla}
\end{figure*}

\subsection{Ablation Study (RQ2)}
To verify the contribution of each module in UTTSI, we construct three ablation variants and evaluate them on all four datasets: \\
\textbf{$\bullet$ w/o-Dual (RD)} removes the dual-signal uncertainty estimation and replaces it with a single logit-only confidence score, disabling the frequency-based prior.\\
\textbf{$\bullet$ w/o-Attr (RA)} removes attribution-guided sampling and replaces it with uniform random feature sampling during path construction. \\
\textbf{$\bullet$ w-Single (WS)} removes multi-path exploration and uses a single inference path per instance, i.e., $K(x)=1$ for all uncertain samples.

Figure \ref{abla} presents the ablation results. The performance degradation upon removing any single component confirms the contribution of each module within our UTTSI.

\textbf{Effect of dual-signal estimation (w/o-Dual, RD).} Replacing the dual-signal estimator with logit-only confidence consistently degrades performance across all datasets and backbones. The root cause is that model logit confidence conflates two distinct sources of uncertainty: epistemic uncertainty arising from sparse feature coverage and aleatoric uncertainty from inherently ambiguous click decisions near the decision boundary. A sample with a near-50\% click probability due to a genuine preference tie will produce a low-confidence logit, yet its feature representations may be perfectly reliable. The frequency-based prior correctly assigns low uncertainty to such samples, preventing wasteful exploration. Conversely, a model may produce high-confidence logits for sparse feature combinations by overfitting to training noise, and the frequency prior is necessary to flag these as high-uncertainty. The consistent gains of the full UTTSI over w/o-Dual confirm that combining both signals is essential for accurate uncertainty discrimination.

\textbf{Effect of attribution-guided sampling (w/o-Attr, RA).} Replacing attribution-weighted sampling with uniform random sampling substantially degrades performance, particularly on datasets with high feature sparsity (KDD12 and Industrial). Under uniform random sampling, the probability of including a high-signal rare feature in any single path is proportional to $1/N$, which can be extremely small when $N$ is large. Since attribution scores directly reflect the base model's assessment of each feature's contribution to the current prediction, using them to up-weight informative features ensures that exploration paths retain the predictive signal of the original input while diversifying away from unreliable feature combinations. The magnitude of degradation for w/o-Attr is larger than for w/o-Dual on the Industrial dataset, suggesting that the quality of individual exploration paths has a greater impact than uncertainty score accuracy in high-dimensional settings.

\textbf{Effect of multi-path aggregation (w-Single, WS).} Even when both dual-signal estimation and attribution-guided sampling are applied correctly, using a single path per uncertain instance underperforms the full multi-path ensemble. Because exploration sampling is stochastic by design, any single path may fail to select the globally optimal feature subset for a given instance. Multi-path sampling addresses this risk by independently exploring $K(x)$ diverse feature configurations and aggregating them via consistency-weighted voting. The consistency weighting further mitigates the risk that any outlier path dominates the final prediction. These results confirm the core design principle of UTTSI: robustness for uncertain instances is derived from the diversity of ensemble exploration, not from finding a single optimal feature configuration.
 
\begin{figure}[t]
  \centering
  \includegraphics[width=\linewidth]{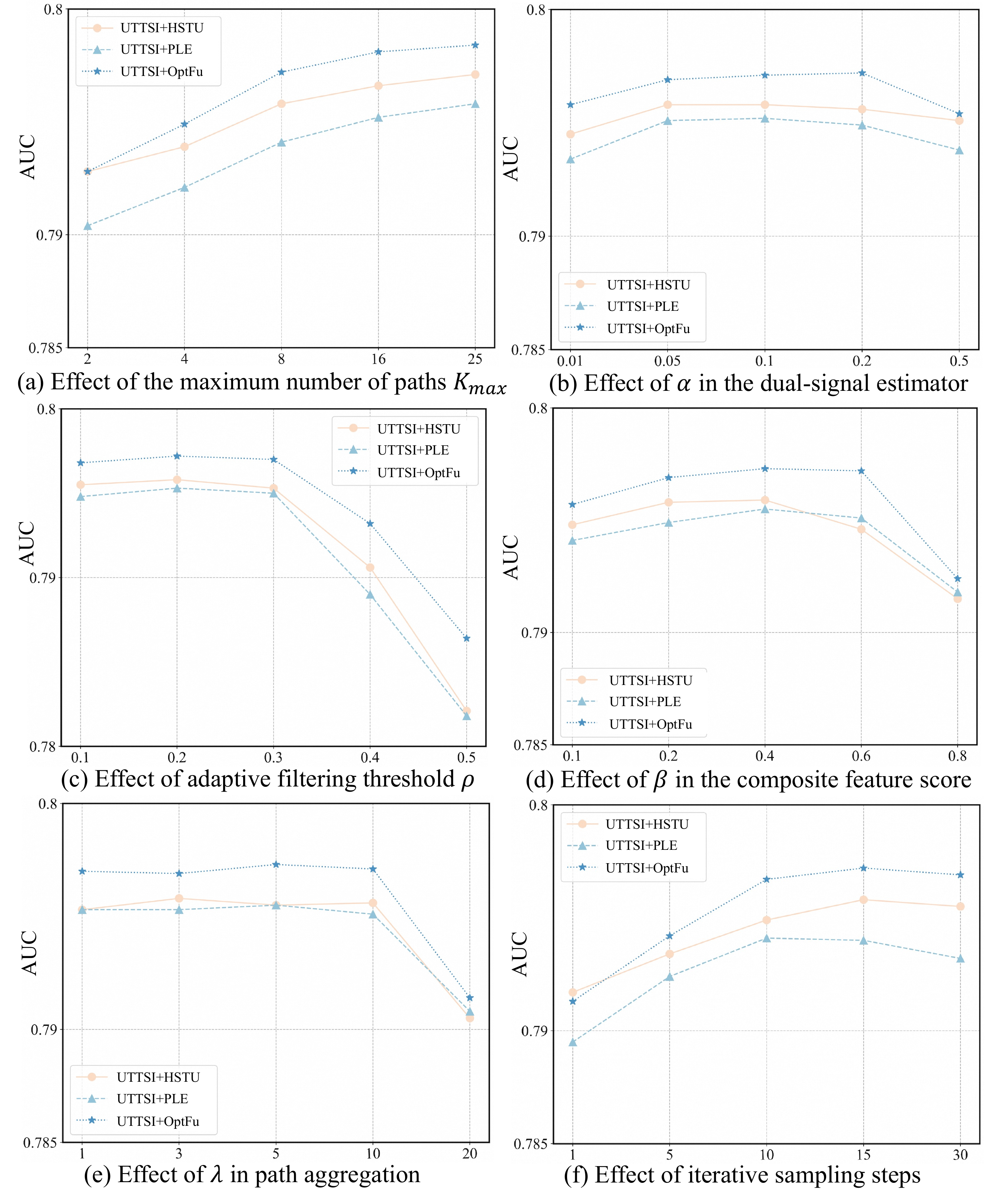}
  \caption{Prediction performance of UTTSI under varying $K_{\max}$, $\alpha$, $\rho$, and $\beta$ on the industrial dataset.}
  \label{hyper}
\end{figure} 

\subsection{Hyperparameter Sensitivity (RQ3)}\label{sec:rq3}
We analyze sensitivity to UTTSI's key hyperparameters on the Industrial dataset, with other hyperparameters fixed at their defaults.

\textbf{Effect of $K_{\max}$.} $K_{\max}$ is the maximum number of exploration paths allocated to any single instance, controlling the upper bound on per-instance inference depth. We vary $K_{\max}$ over $\{2, 4, 8, 16, 25\}$. As shown in Figure \ref{hyper}(a), performance improves as $K_{\max}$ increases from 2 to 8, with diminishing returns beyond $K_{\max}=8$. Crucially, the continuous allocation $K(x)=\lfloor K_{\max} \cdot u(x) \rfloor$ means that the majority of low-uncertainty samples incur $K(x)=0$ and are served directly, so the average number of paths executed per instance remains well below $K_{\max}$. This confirms that UTTSI scales gracefully: increasing $K_{\max}$ improves performance for the hardest instances while leaving efficient and confident samples unaffected.

\textbf{Effect of $\alpha$.} $\alpha$ controls the relative weight of the frequency-based prior versus model logit confidence in the dual-signal uncertainty estimator. We vary $\alpha$ over $\{0.01, 0.05, 0.1, 0.2, 0.5\}$.  As shown in Figure \ref{hyper}(b), performance is stable across a wide range ($0.05$--$0.2$), confirming that the dual-signal uncertainty score is robust to the exact weighting of the frequency prior. Very small $\alpha$ underweights the population-level sparsity signal, reducing uncertainty discrimination for sparse samples, while very large $\alpha$ overrides the model's own confidence signal. We set $\alpha=0.05$ as the default, which provides the best balance between two signals.

\textbf{Effect of adaptive filtering threshold.} $\rho$ is the shared per-field quantile used to compute Stage~1 filtering thresholds $\{\tau^i\}$, governing how aggressively low-scoring features are removed. We vary $\rho$ over $\{0.1, 0.2, 0.3, 0.4, 0.5\}$. Performance is stable across $\rho \in [0.1, 0.3]$, with $\rho=0.2$ achieving the best balance between noise removal and signal preservation. When $\rho=0$ (no filtering), both confident and uncertain instances suffer from unfiltered noisy features, degrading overall AUC. When $\rho$ exceeds 0.4, overly aggressive filtering begins to remove informative rare features, harming the prediction performance in the tail-sample.

\textbf{Effect of $\beta$.} $\beta$ balances frequency reliability and gradient attribution in the composite feature score $w(f^i)$, which is shared by both Stage~1 filtering and Stage~2 path sampling. We vary $\beta$ over $\{0.1, 0.2, 0.4, 0.6, 0.8\}$. Performance is stable across $\beta \in [0.2, 0.6]$, with $\beta=0.4$ optimal. Very small $\beta$ over-weights attribution, retaining noise features with high gradient norms; very large $\beta$ over-weights frequency, discarding informative rare features.

\textbf{Effect of $\lambda$.} $\lambda$ controls the sharpness of consistency weighting in aggregation, determining how strongly outlier paths are down-weighted relative to the ensemble majority. We vary $\lambda$ over $\{1, 3, 5, 10, 20\}$. AUC is insensitive to $\lambda$ within $[1, 10]$ (all values within 0.0005 AUC of the optimum), since consistency weighting primarily suppresses catastrophic outlier paths, and moderate $\lambda$ achieves this effect. Beyond $\lambda=20$, overly sharp weighting approximates winner-take-all selection, losing the benefit of ensembling.

\textbf{Effect of $T$.} $T$ is the number of iterative Bernoulli sampling steps used to construct each feature path in Stage~2, controlling the coverage and diversity of each exploration path. We vary $T$ over $\{1, 5, 10, 20, 30\}$ while fixing all other hyperparameters. Performance improves steadily from $T=1$ to $T=10$, as more steps allow the path construction process to consider a broader set of feature combinations. Beyond $T=10$, gains plateau and performance remains stable up to $T=30$, indicating that the feature space is sufficiently explored within 10 steps for the feature set sizes encountered in our datasets. Very small $T$ (e.g., $T=1$) limits exploration diversity, as paths terminate early with too few features sampled, reducing coverage of high-attribution rare features. We set $T=10$ as the default, which achieves a good balance between exploration quality and computational cost.

Together, these results confirm that UTTSI is robust across all key hyperparameters and generalizes reliably across different hyperparameter settings.

\subsection{Uncertainty Calibration (RQ4)}
A fundamental requirement for any uncertainty-triggered system is that the estimated uncertainty score should correlate with actual prediction error. If the uncertainty estimator is poorly calibrated, high-uncertainty instances may not actually be harder to predict, and routing them to multi-path exploration would yield no benefit. In this subsection, we analyze the calibration of UTTSI's dual-signal uncertainty estimator and its correlation with prediction error.

\textbf{Uncertainty-Error Correlation.} We sort all test instances on the Industrial dataset by their estimated uncertainty score $u(x)$ and partition them into deciles. For each decile, we compute the average squared prediction error $|\sigma(\text{logit}(x)) - y|^2$ of the base model's direct score (without any UTTSI processing). The mean prediction error increases monotonically with estimated uncertainty across all deciles. The Spearman rank correlation between $u(x)$ and squared prediction error is $\rho = 0.91$ on the Industrial dataset, confirming that the dual-signal estimator reliably identifies instances where the base model is likely to err. By contrast, using logit confidence alone (i.e., the UTTSI-w/o-Dual variant) yields a substantially lower Spearman correlation of $\rho = 0.76$, consistent with the ablation results in RQ2 and validating the necessity of the frequency-based prior for uncertainty discrimination. We further verify this calibration property on the Criteo and Avazu datasets, obtaining Spearman correlations of $\rho = 0.88$ and $\rho = 0.86$ respectively for the dual-signal estimator, versus $\rho = 0.71$ and $\rho = 0.73$ for logit-only confidence. The consistent calibration superiority across three datasets with distinct feature structures confirms that the dual-signal design generalizes beyond the industrial setting.

\textbf{Performance Across Uncertainty Strata.} To further understand where UTTSI's gains originate, we compare the AUC of the base model versus OptFu \cite{dag}+UTTSI separately on low-uncertainty (bottom 30\% by $u(x)$), medium-uncertainty (middle 40\%), and high-uncertainty (top 30\%) subgroups of the Industrial test set. The results are summarized in Table \ref{calib_table}.

\begin{table}[t]
\small
\centering
\begin{tabular}{l|cc|c}
\hline
Stratum & OptFu & UTTSI & $\Delta$ \\
\hline
Low ($u(x) < 0.3$) & 0.8241 & 0.8254 & +0.0013 \\
Medium ($0.3 \le u(x) < 0.7$) & 0.7985 & 0.8012 & +0.0027 \\
High ($u(x) \ge 0.7$) & 0.7621 & 0.7709 & +0.0088 \\
\hline
Overall & 0.7933 & 0.7979 & +0.0046 \\
\hline
\end{tabular}
\caption{AUC by uncertainty stratum on the Industrial dataset (OptFu backbone). UTTSI allocates the most exploration budget to the high-uncertainty stratum, where gains are largest.}
\label{calib_table}
\end{table}

The results reveal a clear pattern: UTTSI provides a consistent positive gain across all uncertainty strata. Even for low-uncertainty samples ($K(x)=0$), the adaptive feature filtering removes features with unreliable representations and yields a modest but consistent AUC improvement of $+0.0013$ ($p < 0.05$, paired permutation test), confirming that the gain is not spurious. Medium-uncertainty samples benefit more substantially, and high-uncertainty samples achieve the largest gain of $+0.0088$ thanks to multi-path exploration. This stratified pattern confirms that UTTSI's design is coherent: feature filtering benefits all instances, while additional exploration paths are reserved for the hardest cases: it invests additional inference resources exactly where the base model is weakest, without degrading the overall efficiency of the system. These results confirm that both uncertainty calibration and multi-path exploration are necessary: the former ensures budget is not wasted on easy instances, and the latter converts that budget into the prediction performance gains on hard ones.

\subsection{Online A/B Testing Results}
To validate real-world efficacy, we deployed UTTSI in a seven-day online A/B test on a large-scale e-commerce platform from April 15 to 21, 2026. Compared against a production baseline model with a PEPNet-like architecture \cite{pepnet}, UTTSI achieved a 5.3\% relative improvement in click-through rate ($p < 0.01$, two-sided $z$-test on daily CTR metrics across the seven-day window, with randomized traffic splitting at the user level). This CTR improvement is substantially larger than the offline AUC delta because online CTR measures the top-ranked items' actual click yield, where even small ranking improvements at the head of the recommendation list translate into large relative CTR gains. This statistically significant gain in a live production environment provides compelling evidence of the framework's practical value.

\textbf{Computational overhead analysis.} To quantify the practical overhead introduced by UTTSI, we measure the average number of model calls per instance on the Industrial dataset. With $K_{\max}=8$, $\alpha=0.05$, and $\rho=0.2$, the empirical distribution of $K(x)$ reveals that approximately 62\% of samples receive $K(x)=0$ (adaptive feature filtering only), 21\% receive $K(x) \in \{1, 2\}$, and the remaining 17\% receive $K(x) \geq 3$. Every instance requires one base model call (forward pass with gradient attribution for uncertainty estimation) and one refined forward pass on $\hat{\textbf{F}}$; uncertain instances additionally receive $K(x)$ exploration forward passes. This amounts to an average of approximately 2.8 model calls per instance, representing a $2.8\times$ overhead relative to the base model. The refinement pass uses the same model architecture with a masked input, so its cost is comparable to the base forward pass. Given that all exploration paths are fully independent and parallelizable, this overhead translates directly to additional parallel compute capacity rather than increased wall-clock latency in a horizontally scaled serving system. When the serving infrastructure supports at most $K_{\max}$ parallel scoring workers, UTTSI's worst-case latency equals one base model forward pass, which is equivalent to the original serving latency. This makes UTTSI particularly well-suited to industrial systems where parallel compute is readily available but strict latency SLAs must be maintained.

\textbf{Frequency index maintenance.} In production, the frequency index is not static: new items and users are introduced daily, and existing feature values accumulate additional impressions over time. To keep the frequency prior current, we maintain the Count-Min Sketch structure with daily incremental updates: at the end of each day, the previous day's impression logs are merged into the sketch by replaying the hash-and-increment operations for each observed feature value. Because the Count-Min Sketch supports monotone increments and its minimum-across-tables estimator remains valid after incremental updates, no full recomputation is required; only the new day's records need to be processed, making the update cost proportional to the daily impression volume rather than the full historical corpus. The normalized counts $\tilde{s}_{freq}(f^i)$ and the per-field filtering thresholds $\{\tau^i\}$ are recomputed from the updated sketch after each nightly merge and pushed to the serving layer before the next day's traffic, ensuring that uncertainty estimates and filtering decisions reflect up-to-date feature coverage. During the A/B test window, this daily refresh cycle maintained stable calibration: the Spearman correlation between $u(x)$ and observed prediction error did not degrade across 7 days, confirming that incremental updates are sufficient to track the distribution of feature frequencies.

\section{Conclusions}
In this paper, we proposed a model-agnostic training-free Uncertainty-Triggered Test-Time Selective Inference framework (UTTSI) that addresses a fundamental limitation overlooked by training-centric CTR research: test-time prediction reliability varies substantially across instances due to feature sparsity, yet no existing mechanism allocates computation selectively to compensate for this variability. UTTSI resolves this through three coordinated steps: (1) dual-signal uncertainty estimation that combines model logit confidence with a frequency prior to produce a per-instance reliability score; (2) adaptive feature filtering applied to every instance to remove embeddings with insufficient training support; and (3) selective multi-path exploration for uncertain instances, whose diverse predictions are aggregated via consistency-weighted ensembling. The framework is plug-and-play, requiring no retraining or architectural changes, and is compatible with any CTR backbone. Experiments across four datasets and a seven-day online A/B test confirm consistent improvements, with $2.8\times$ average computational overhead (one refinement pass for all instances plus selective multi-path exploration for uncertain ones) manageable within industrial serving constraints. These results establish selective inference at test time as a principled and complementary direction to training-phase advances for CTR prediction.

\end{document}